# Mapping bathymetry of inland waterbodies on the North Slope of Alaska with Landsat using Random Forest

Carroll, M.L.[1,a], Wooten, M.R.[2,3], Simpson, C.E.[4], Spradlin, C.S.[1,5], Frost, M.J.[1,5], Blanco-Rojas, M.[1], Williams, Z.W.[1,5], Caraballo-Vega, J. A.[1], Neigh, C.S.R.[2]

1: NASA Data Science Group, Goddard Space Flight Center, 8800 Greenbelt Rd. mail code 606.3 Greenbelt, MD 20771, USA

2: NASA Biospheric Sciences Laboratory, Goddard Space Flight Center, 8800 Greenbelt Rd. mail code 618 Greenbelt, MD 20771, USA

3: Science Systems and Applications Incorporated, 10210 Greenbelt Rd Suite 600 Lanham, MD 20706, USA

4: Department of Geography, University of Colorado Boulder, Boulder, Colorado, 80309, USA

5: ASRC Federal Goddard Space Flight Center, 8800 Greenbelt Rd. mail code 606.3 Greenbelt, MD 20771, USA

a: corresponding author mark.carroll@nasa.gov

## Abstract

The North Slope of Alaska is dominated by small waterbodies that provide critical ecosystem services for local population and wildlife.  Detailed information on the depth of the waterbodies is scarce due to the challenges with collecting such information.  In this work we trained a machine learning (Random Forest Regressor) model to predict depth from multispectral Landsat data in waterbodies across the North Slope of Alaska.  The greatest challenge is the scarcity of in situ data, which is expensive and difficult to obtain, to train the model.  We overcame this challenge by using modeled depth predictions from a prior study as synthetic training data to provide a more diverse training data pool for the Random Forest.  The final Random Forest model was more robust than models trained directly on the in situ data and when applied to 208 Landsat 8 scenes from 2016 – 2018 yielded a map with an overall $r^2$ value of 0.76 on validation.

The final map has been made available through the Oak Ridge National Laboratory Distribute Active Archive Center (ORNL-DAAC). This map represents a first of its kind regional assessment of waterbody depth with per pixel estimates of depth for the entire North Slope of Alaska.

## Plain Language Summary

The North Slope of Alaska is a complex mosaic characterized by minimal elevation change and numerous small water bodies, primarily lakes, ponds, and wetlands. These water bodies are critical to the subsistence lifestyle of inhabitants and wildlife however little is known about their function. One elusive function characteristic is the depth of the water bodies. Using machine learning we modeled not only the average depth but also the depth profiles of each waterbody based on the optical properties of the water using satellite data from the Landsat missions. This enables scientists to better understand the behavior of the water bodies within a modeling context so they can understand how climate change may affect the region.

## 1. Introduction

The landscape in the North American Arctic tundra is characterized by low vegetation—mainly grasses, sedges and moss—with numerous small waterbodies. In recent years researchers have mapped the location and extent of these waterbodies (Andresen and Lougheed, 2015; Carroll et al., 2016; Carroll and Loboda, 2017; Muster et al., 2017; Pekel et al., 2016; Rover et al., 2012), however these metrics alone are insufficient to describe how these waterbodies function. Other essential components of function including extent changes and depth have not been adequately addressed. Recent work has focused on quantifying surface area change over time (Carroll et al., 2011b; Carroll and Loboda, 2018; Chen et al., 2014; Cooley et al., 2017;

Smith et al., 2005; Smol and Douglas, 2007). However, very little has been done to quantify depth in Arctic lakes (Duguay and Lafleur, 2003; Grunblatt and Atwood, 2014; Jeffries et al., 1996; Kozlenko and Jeffries, 2000).

Depth is a critical factor for measuring volumetric storage of water, modeling limnology, and quantifying the ecological services created by that waterbody. Inclusion of a lake model into climate models requires limnology (Bonan, 1995; Subin et al., 2012). The depth of a lake is a determining factor in whether ice forms to the bottom of the lake in winter. Lakes that do not freeze to the bottom represent important resources, providing overwintering habitat for fish (Berkes and Jolly, 2001), providing freshwater sources for indigenous people (Alessa et al., 2008; Eichelberger, 2018; White et al., 2007), and providing a resource for industrial use (such as ice road construction) (Jones et al., 2009).

The arctic contains millions of small lakes, ponds and wetlands (Carroll et al., 2016; Downing et al., 2006; Pekel et al., 2016). Physically measuring the bathymetry of even a representative sample of these lakes is infeasible. It is therefore necessary to find methods to estimate or to measure the depth using remote sensing methods. Several remote sensing systems can be used to measure depth in waterbodies including Light Detection and Ranging (LiDAR) (Moyles et al., 2005; Paine et al., 2013; Saylam et al., 2017), Synthetic Aperture Radar (SAR) (Duguay and Lafleur, 2003; Grunblatt and Atwood, 2014; Jeffries et al., 1996; Kozlenko and Jeffries, 2000), and multi-spectral remote sensing as far back as the early 1980's (Lyzenga, 1981). LiDAR remote sensing is limited because 1) the space-borne LiDAR GEDI and ICESat-2 are not designed to measure bathymetry, though ICESat-2 does show some potential in this area (Forfinski-Sarkozi and Parrish, 2016; Yang et al., 2023), and 2) because the density of observations from LiDAR is insufficient for complete mapping of water bodies. SAR methods are often applied to winter data and measure the thickness of the ice rather than the actual depth of the water, therefore providing only a coarse understanding of water body depth in deeper water conditions where lakes do not fully freeze (Grunblatt and Atwood, 2014; Jeffries et al., 1996; Kozlenko and Jeffries, 2000). Methods for measuring depth of water from multi-spectral remote sensing usually involve a linear regression on one or more spectral bands, or ratios of bands (Duguay and Lafleur, 2003; Lyzenga, 1981; Stumpf et al., 2003), though some recent attempts have been made at using machine learning (Random Forest and Stereoscopy) methods.

Machine learning algorithms are a group of mathematical models that enable computers to learn patterns in data without explicit programming. Image classification is a subfield of machine learning with the capacity to understand visual patterns by assigning class labels to image pixels. In the field of remote sensing, a typical image classification workflow involves collecting field samples and attributing them with features/predictors extracted from spectral bands or derivatives of satellite imagery (Breiman, 2001; Breiman et al., 1984). These algorithms have been used extensively for land cover classification (Carroll et al., 2011a, 2009; Chang et al., 2007; DeVries et al., 2017; Hansen et al., 2008, 2002). Conceptually, the problem of calculating water depth from remotely sensed data is well suited to machine learning, as the dataset is comprised of field measurements (depth readings from sonar or other methods) which can be related to a set of features (spectral bands from imagery) that were not collected in situ but nevertheless have relevance for predicting the model target (depth). The relationship of depth to spectral response varies based on the composition of the water (turbidity, chlorophyll content, etc.) and bottom color/type (soft/hard, bright/dark, etc.). The machine learning algorithm can learn from the training data and provide novel inferences from the spectral data that may not be obvious to a human interpreter.

Recent work has been performed using various machine learning techniques to predict depth using visible and Near infrared imagery from Worldview, the Landsat suite or Sentinel 2 imagery (Chen et al., 2022; Deidda and Sanna, 2012; Manessa et al., 2016; Merchant, 2023; Yang et al., 2022). In each case, these studies use a set of in situ measurements to train a model that can generalize across an entire image or across multiple images. More recently, researchers used a Radiative Transfer model to use the remote sensing physics to predict depth in near shore coastal ocean with some success (Xu et al., 2023). The limiting factor in the previous work has been the sparse in situ measurements used to train the model. To overcome this problem in other machine learning contexts synthetic data augmented the data pool to increase the robustness of the data or to preserve the physical measurements (Hittmeir et al., 2019). Synthetic data is any data from a simulator or model that represents features of interest but were not directly measured. In our case, Simpson et al,. (2021) used linear regression models to make prior depth predictions on individual lakes that could serve as synthetic data in training a machine learning model.

Our objective is to demonstrate the utility of machine learning in general and specifically the use of synthetic data in training a Random Forest machine learning model for predicting depth in water bodies on the North Slope of Alaska. We designed three experiments to determine how best to use the available data to generate a generalized Random Forest model for estimating water body depth on the North Slope of Alaska:

1) Use only training from 17 lakes and a single Landsat scene where previous work using linear models has been done and quantify results (Simpson et al., 2021);

2) Collected additional training data from Landsat 8 using additional dates not adjacent to the date of in situ collection to train a more robust and generalized model;

3) Use maps produced in prior work (Simpson et al., 2021) as synthetic data to gather training from Landsat 8 to improve the diversity of samples in the training pool.

The overall goal is to define a method that can be used to make a map of depth for lakes on the North Slope of Alaska. This study is the first of its kind to attempt to make a per pixel map of depth for water bodies on the entire North Slope of Alaska.

## 2. Study area

The study area is on the North Slope of Alaska, USA (figure 1). This area is in the Arctic Tundra ecoregion (Olson et al., 2001) characterized by low topographic elevation (generally < 150 m), short vegetation and moss, with minimal precipitation (15 – 25 cm per year). This area was chosen due to existing depth measurements collected in situ in 2017 (Simpson and Arp, 2017).

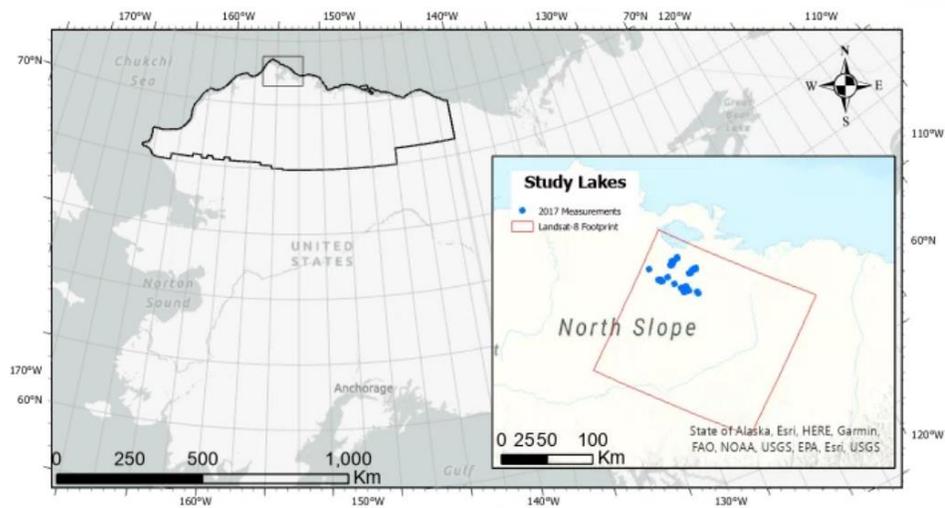

*Figure 1 Study area in North Slope of Alaska, USA. Points represent lakes where depth measurements were collected in Simpson et al. 2021. Red outline shows the footprint of the Landsat scene that was used in experiment 1 of this study.*

## 3. Methods

The Random Forest model requires a set of training data (dependent variable) and a set of predictors (independent variables) to generate a model to predict the depth from the predictors. In previous work, the coauthors collected depth information using sonar mounted on the pontoon of a sea plane (Simpson and Arp, 2017). The sonar measured and recorded depth frequently, resulting in thousands of measurements. These points were spatially aggregated to represent Landsat 30 m pixels by calculating mean depth per 30 m pixel using all depth points that fall within that pixel (see Simpson et al. 2021 for full description of how points were handled).

All Landsat Collection 2 Surface Reflectance data were acquired from the USGS ESPA on demand interface (https://www.usgs.gov/media/images/espa-demand-interface). At high latitudes Landsat paths overlap more than equatorial latitudes, hence many paths were used to overlap a relatively small geographic area which allowed us to maximize the total observations of each lake. All data were processed on the Explore/ADAPT system at the NASA Center for Climate Simulation (https://www.nccs.nasa.gov/systems).

### 3.1. Experiment 1: Random Forest compared to linear model

In previous work the coauthors used depth points collected in summer 2017 (Simpson and Arp, 2017) with a single path row from Landsat 8 (path 76 row 011 08/05/2016) to generate depth and volume estimates for 17 sampled lakes using a linear regression modeling approach (Stumpf et al., 2003). The results showed good agreement with the field measurements on individual lakes but the model had to be trained for each lake and was thus not transferable or extensible to other lakes in the region. Here we used the same input depth data and input Landsat data in a Random Forest model to determine if we could get a single model to represent all lakes while retaining comparable accuracy. We used the Random Forest Regressor in Python's scikit-learn module with hyperparameters max_feat = sqrt, number of trees = 100 and an 80/20 split of training/test data. The Random Forest models were trained with 668 samples (the number of points that remained after reducing the 13,735-point dataset to 30-m resolution to match Landsat spatial resolution) acquired across 17 lakes and spanning a depth range of 0.2 - 21.0 m. The Random Forest model was then applied to Landsat surface reflectance data from 2017 for direct comparison with previous results (Simpson et al., 2021).

The maps of the 17 lakes from Simpson et al. 2021 and from the Random Forest method were clipped to the extent of lakes based on the ABoVE Water Maps (Carroll et al., 2016) to ensure compatibility in spatial extent between the results. Outputs of each lake were compared to assess any differences between the two methods.

### 3.2. Experiment 2: Train and apply Random Forest method on time series of Landsat data

On the North Slope of Alaska most lakes are "closed," i.e., they have no major outlet. These lakes will therefore have a reasonably stable depth profile from year to year unless there is a significant event that causes drainage (Jones et al., 2009). We inspected a time series of Landsat data for the lakes with available depth measurements and determined the surface water extent to be generally consistent through time. We can assume that where the surface extent has not significantly changed, the depth also does not change significantly. As a result, measurements of depth from one time period is presumed to be representative of the depth at other times even though the exact depth will be slightly different.

The primary limitation of the training data (Simpson and Arp, 2017) used in experiment 1 is the small number (668) of points with which to train after aggregation to Landsat spatial resolution.. We compiled all Landsat 8 surface reflectance scenes (path 074 – 080 row 010 – 011) from 2017 (the year of depth data collection) that were free of ice and clouds (determined using Landsat's quality flags, nominally July and August) and covered lakes in experiment 1. For each image in this collection, we extracted spectral band values at each of the in situ depth sample locations. These samples were then assigned the same depth values as the original measurements in experiment 1. This effectively allowed us to expand the range of spectral values associated with various depths. The expansion of training data was necessary to account for natural atmospheric and image quality variations in the Landsat data and facilitating better model generalization. This expanded our number of training samples from 668 to 24,233 which greatly enhanced our ability to generalize across both space and time with the Random Forest model. We use the same hyperparameters as experiment 1 for consistency in methodology.

### 3.3. Experiment 3: Random Forest model using linear regression results as "synthetic" training data and a time series of inputs

Though we expanded the number of training points in experiment 2 we still had a very small training data pool for the expansive study area. Time and cost constraints prevented additional in situ sonar-based data collection to supplement our training data. The only other source of information we had available were the maps generated in previous work (Simpson et al., 2021). If we consider that these maps are published (Simpson and Arp, 2017) and have been validated with support from a peer reviewed publication, then we can use the maps themselves as "synthetic" training data. Synthetic data is any data from a simulator or model that represents features of interest but were not directly measured. We sampled the time series of Landsat 8 data underneath the mask generated from Simpson's maps. This yielded a more robust training data set, more than 1 million points, with a much wider range of depth values. We use the same hyperparameters as experiment 1 for consistency in methodology.

## 4. Results

### 4.1. Experiment 1: Linear regression vs Random Forest model

The Random Forest model trained on just the points from the 17 lakes produced outputs with similar patterns to the results from the linear regression model. Model statistics for the Random Forest models for all of the experiments are shown in table 1. A scatterplot of the linear regression results for 17 lakes with the Random Forest results for the same lakes shows good agreement when the depth is less than 10 m and an increasing "shallow" bias with greater depths (figure 2).

Table 1 Model statistics for all three experiments.

| Experiment | Training $r^2$ | Validation $r^2$ | Mean Absolute Error | Out of Bag (OOB) Accuracy |
|---|---|---|---|---|
| 1 | 0.88 | 0.31 | 1.06 | 0.13 |
| 2 | 0.93 | 0.55 | 0.35 | 0.56 |
| 3 | 0.98 | 0.85 | 0.84 | 0.52 |

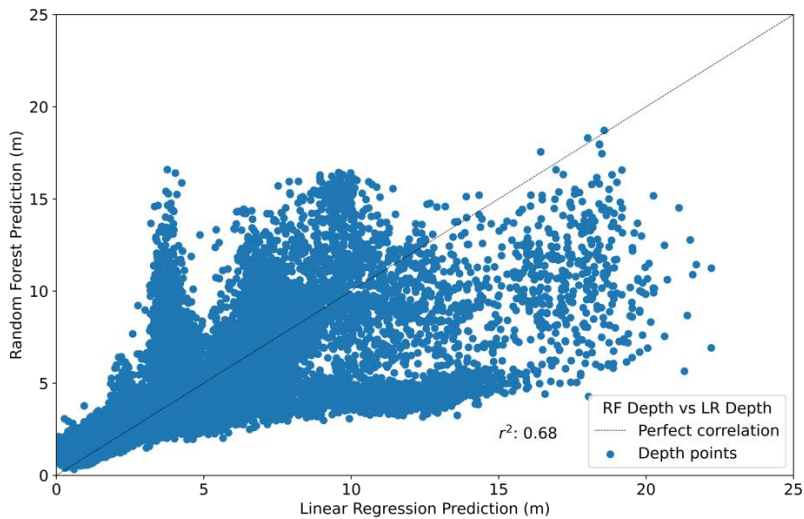

*Figure 2 Linear regression results for the 17 lakes in the North Slope compared to the locally trained Random Forest model.*

This result is reasonable when considering that the linear regression models are specific to each lake (i.e. one model for each lake, no generalization) and the Random Forest is a single model for the entire scene using information from all 17 lakes. The differences can be seen spatially in figure 3 below which shows the difference between the Linear regression results and the Random Forest results. Positive differences indicate that the Linear regression predicted a deeper depth.

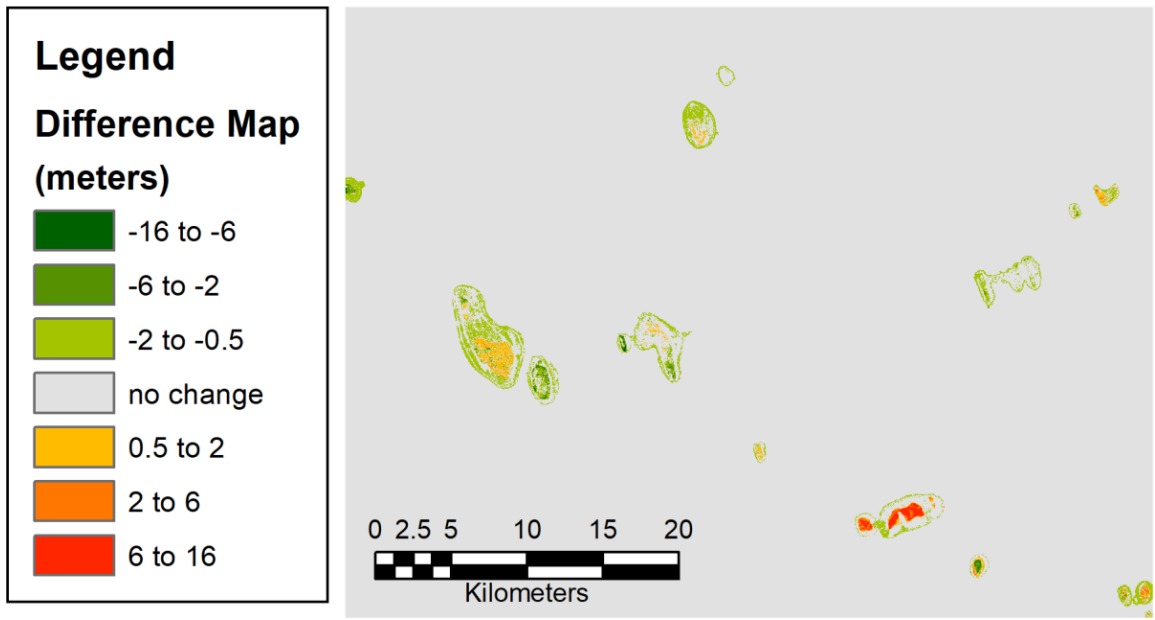

*Figure 3 Difference map showing how the linear regression generated map compares to the Random Forest generated map. The green colors indicate that the RF map has deeper values and the red colors show the linear map has deeper values.*

The 17 lakes were represented by slightly more than 300,000 pixels in total. Overall there was good agreement between the two methods with 238,480 pixels (~79%) having a difference between -0.1 m and 0.1 m. Over 95% of the differences were within +/- 2 meters. One outlier lake had a consistently deeper prediction in the linear regression model (the lake with large red area in the lower right of figure 3). This lake was optically different from the others and visual interpretation suggests that the lake appears to have a different substrate (i.e. less sandy/bright) which may be contributing to the difference.

### 4.2. Experiment 2: Random Forest expanded points and time series of inputs

The generalized Random Forest model described in the methods section under experiment 2 performed reasonably well (table 1). As expected, increasing the number of training points by expanding the input Landsat scenes resulted in improvements to all model statistics. This model was applied to a total of 31 Landsat 8 Surface Reflectance scenes across two years spanning 2016 – 2018. Results from all scenes for a given year were combined to generate a single "composite" result for each year. The mean, median and max depth, total observation count, and standard deviation were recorded for each pixel. We initially produced maps for each year, however we found that the observation count per pixel varied wildly from year to year (mostly due to cloud cover obscuring observations) making it difficult to compare the results. Ultimately, we created a single summary map combining the data from all years to produce median and maximum depth, total observation count, and standard deviation per pixel.

A scatterplot, figure 4, shows a sharp cutoff in predicted depths greater than 4 m, a strong bias to towards shallow depths. This suggests that even though we expanded our training data we have not yet fully captured the range of depths with a sufficient representation of the spectral data from Landsat.

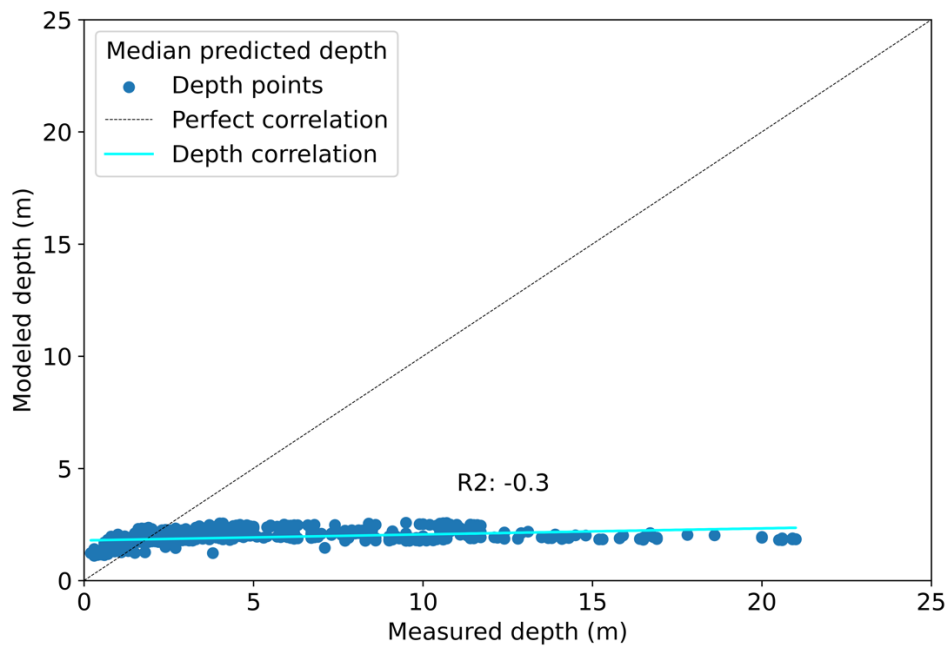

*Figure 4 Scatterplot showing results from the generalized Random Forest model compared to the observations from in situ measurements.*

### 4.3. Experiment 3: Random Forest trained with synthetic data

The Random Forest model trained with synthetic data yielded reasonable statistics (table 1). This model was applied to the 208 Landsat 8 scenes that overlapped the lakes with synthetic data spanning the range from 2016 – 2018. The results were combined to produce median depth, max depth, total observation count and standard deviation similarly to the description above under the results for experiment 2. An analysis of the 668 observations from experiment 1 are shown in figure 5. There is still a clear shallow bias in the modeled depth in the median depth composite however the bias is much improved compared to experiment 2. The standard deviation increases with depth, unsurprisingly, however most values show a standard deviation < 3 m.

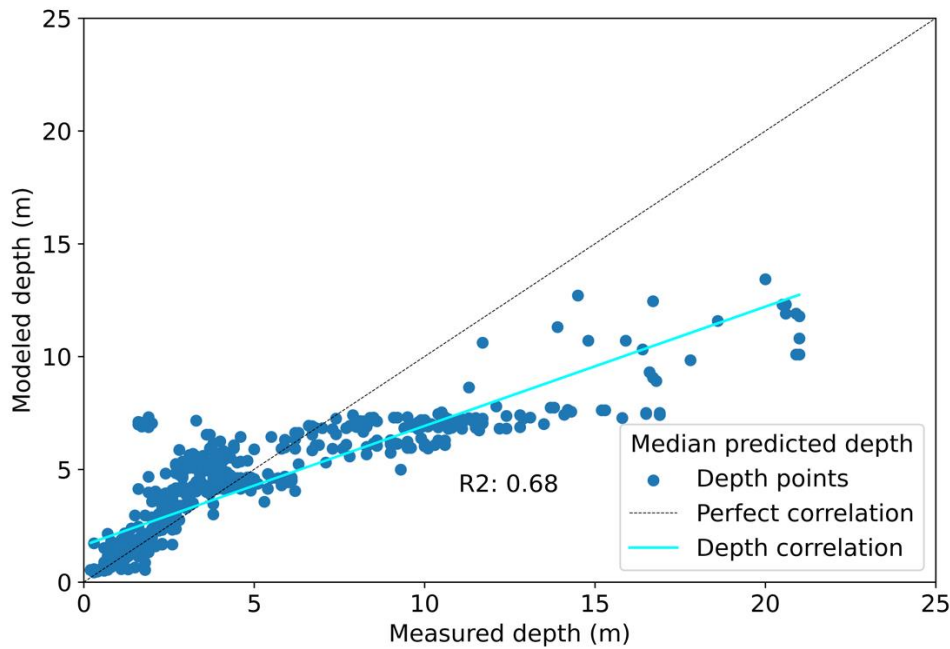

*Figure 5 Scatterplot showing the comparison of the results from the Random Forest model trained with synthetic data compared to the measured in situ observations.*

### 4.4. Final map generation

We generated a map of the full North Slope region of Alaska by applying the generalized model using synthetic training from experiment 3 (figure 6). In total 208 Landsat 8 Surface Reflectance scenes, paths 063-084 and rows 010-013, are available covering the date range 2016 – 2018, ice free months July and August. The model was applied to all of these scenes to generate a time series of depth measurements. To improve confidence and reduce inter-scene variability the predictions were combined into annual and rolling multi-annual representations of depth. Multiple different combinatorial methods were tested before settling on a final method of using three years and taking the median depth value. For the final map the date range of input values was reduced to within plus or minus one year from the original collection of in situ depths at the 17 test lakes in 2017 (i.e. 2016 – 2018). This range was determined empirically by using different combinations of inputs from 2013 – 2022 and performing validation against the 668 observations.

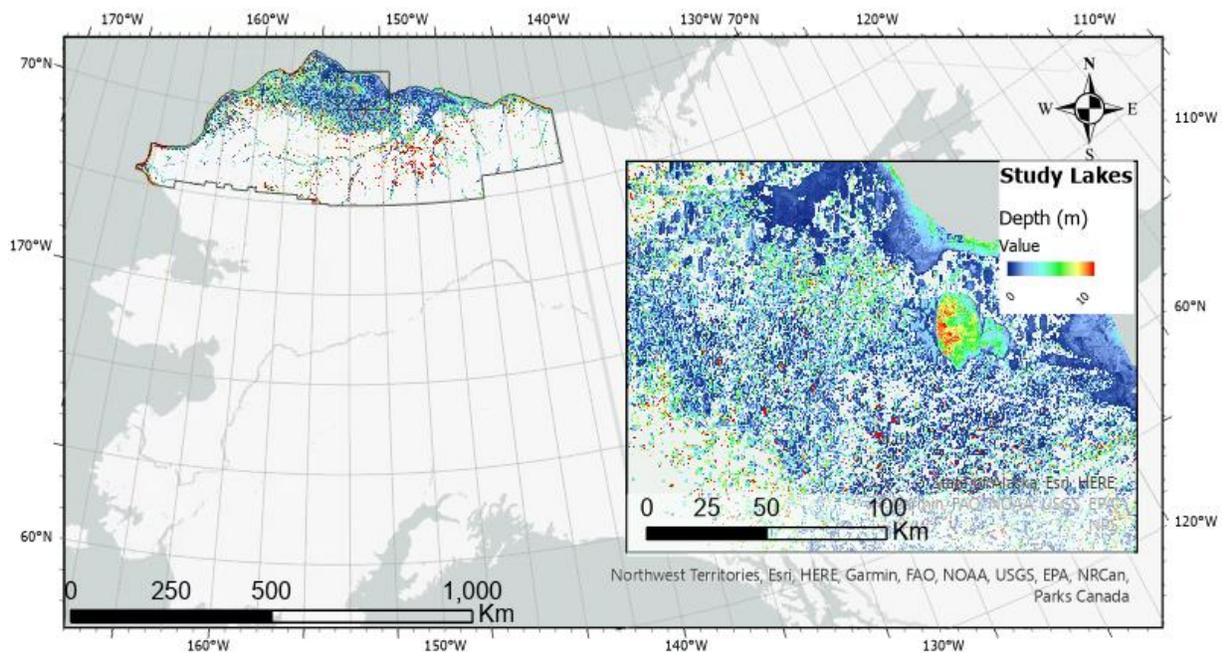

*Figure 6 Study area in North Slope of Alaska, USA. Colors represent the predicted median pixel depth in meters of the lakes in the study area.*

We clipped the resulting map to fit the North Slope Borough boundaries (North Slope Borough, Alaska., 2005). Lastly, the maps were projected and subset into the projection and grid used for the NASA Terrestrial Ecology program's "Arctic and Boreal Vulnerability Experiment" (ABoVE) project. The final map has been delivered to the Oak Ridge National Laboratory Distributed Active Archive Center (ORNL-DAAC) for distribution as a data set in support of the ABoVE project. These data are freely available to any interested parties (https://doi.org/10.3334/ORNLDAAC/2243).

The predicted depths were validated against 668 in situ sonar depth measurements used in experiments 1 – 3 for lakes in the study domain but not directly used to train the final Random Forest model, yielding an overall $r^2$ value of 0.76. This value represents how closely our predicted depths match actual measurements but does not necessarily indicate the accuracy of the predictions. Accuracy is better assessed by evaluating the ratio between correctly predicted depths and the total number of predictions for a given depth range. In our case, this value fell between 82 – 94%.

Since the depth measurements do not follow a normal distribution, we opted to use the normalized median absolute deviation (NMAD) as the measure of variability instead of the standard deviation, given that the NMAD is less affected by potential outliers. The points used for validation each had between 1 and 16 individual depth predictions (count), and their NMAD fell between 0 and 6.7 m. It is important to note, however, that only 12.3% of our predictions had a NMAD of more than 2 m, and for 53% of our validated predictions this value fell below 0.5 m. This implies a high degree of agreement between our predictions and the in situ sonar depth measurements we validated against. Further investigation with more diverse in situ data is needed to understand why the predicted depths deviate as we get farther from the original collection date.

## 5. Discussion

Researchers have had an interest in improving our understanding of the depth profiles of remote water bodies for several decades. This information can be used to identify ecological niches, identify human hazards, and quantify available resources. The typical method of linear regression per lake has proven successful in numerous publications on both coastal waters and lakes. The main limitation of the method is its lack of extensibility to additional water bodies in both space and time. In the first part of our research, we performed a direct comparison of results from a linear regression method and a machine learning method (Random Forest model). The results compare favorably both visually and quantitatively (figures 2 and 3). The favorable comparison between the single scene Linear Regression and single scene Random Forest results motivated the second half of the work in which our goal was to generalize further with additional training from alternate dates of Landsat data.

Generalizability is one of the strengths of machine learning models including Random Forest, hence it is not surprising that we were able to attain good results in creating a model from the additional training data. We demonstrated this in two ways. First, we use multiple input dates of Landsat data to extract spectra at the 668 point locations with in situ depth measurements which greatly expands our input training data. Second, we use the maps generated by the linear regression method as synthetic training data to diversify the set of depth values to train the model. Ideally, we would have a much larger pool of in situ observations, unfortunately it is cost prohibitive to collect this type of data in remote regions such as the North Slope of Alaska. The Random Forest model trained with the synthetic data performed best overall and was used to generate the map of the full North Slope of Alaska.

The map generated from this model, while it should be considered preliminary, provides more information on remote lakes in Alaska than has been previously available. Most researchers who need depth information for their studies rely on an estimate of average depth based on the size (defined by the mappable surface area). The Globathy dataset provides depth information for all lakes shown in the HydroLakes dataset (Khazaei et al., 2022; Messager et al., 2016). An evaluation of the Globathy data reveals that on the North Slope of Alaska the depths are generated by simple interpolation assuming that the center of the water body is the deepest point and there is a uniform descent to the central point. While this can provide a crude estimate for volume calculations it provides only a single average value for the entire lake. The Random Forest model provides values for each pixel in the lake which is necessary when trying to understand environmental phenomena such as wetting and drying around these lakes which can affect Carbon release to the atmosphere.

Work by (Chen et al., 2022) also used Random Forest and the in situ depth from (Simpson and Arp, 2017), however they were only able to produce a map for a small area and limited their work to single images from the year of data collection (2017). By using synthetic data for training and holding the in situ data back for validation we were able to greatly expand our training pool, build a more robust model, and stretch the application of the model to additional years. Lastly, by processing all available Landsat data for the region we were able to generate multiple observations per pixel. This provided a measure of repeatability that we captured in the NMAD statistic to give users of the map a way to identify the highest quality depth predictions and isolate lesser quality predictions.

Future work should include acquiring additional in situ training data for the numerous smaller lakes, additional evaluation to understand why the predictions were weaker as we deviate from the collection date of the in situ data, and potential expansion to additional Landsat sensors or the Harmonized Landsat Sentinel data. Additionally, a newly published study shows Collection 2 Landsat surface reflectance over inland water bodies has biases in some bands used in this study (Maciel et al., 2023). The new aquatic surface reflectance should be investigated in future studies to determine if there is any impact on depth measurements calculated from standard surface reflectance.

## 6. Conclusions

This study demonstrates the capacity of a machine learning model to predict water body depth at scale, and ultimately, to produce a generalized map of water body depth across the North Slope of Alaska.  Traditional methods like linear regression are not portable between lakes so this method provides a way to generate regional maps.  Sparseness of training data remains a significant challenge due to the expense of collecting training data in remote regions.  We addressed this challenge by leveraging "synthetic" training data (generated using the traditional linear regression methods per lake) to provide a diverse training dataset for a Random Forest model.  The models yielded strong statistics with the final model having training $r^2$ of 0.98, validation $r^2$ of 0.85 and Mean Absolute Error less than 1 m.  The preliminary map shows an overall $r^2$ of 0.76 compared to in situ data.  This map has been made publicly available through the ORNL-DAAC.

## Acknowledgements

Computing resources supporting this work were provided by the NASA High-End Computing (HEC) Program through the NASA Center for Climate Simulation (NCCS) at Goddard Space Flight Center.

## Open Research

All Landsat data used as inputs in this work are freely available from the US Geological Survey Earth Explorer system (https://earthexplorer.usgs.gov/).  The sonar measurements used for training were cited in the text and are freely available https://doi.org/10.18739/A2JD4PP1H.  The outputs from this work are freely available from the Oak Ridge National Laboratory Distribute Active Archive Center (ORNL-DAAC) https://daac.ornl.gov/cgi-bin/dsviewer.pl?ds_id=2243.  All processing was performed using open-source Python packages including Pytorch and Geospatial Data Abstraction Library (GDAL).